\pdfoutput=1
\documentclass[sigconf]{aamas} 

\usepackage{balance} %

\usepackage{algorithm}
\usepackage[noend]{algpseudocode}

\usepackage{subcaption}

\usepackage{pgfplots}
\pgfplotsset{compat=1.12}

\usepackage{makecell}

\usepackage[textsize=tiny, textwidth=1.5cm, colorinlistoftodos, disable]{todonotes}
\newcommand{\Aron}[1]{\todo[color=yellow!10, linecolor=yellow!50!black]{\textbf{Aron:} #1}}

\newcommand{\iAron}[1]{\todo[color=yellow!10, inline]{\textbf{Aron:} #1}}

\newcommand{\ad}[1]{\todo[color=green!10]{\textbf{AD:} #1}}

\usepackage[capitalize, noabbrev]{cleveref}

\setcopyright{none}
\acmConference[AAMAS '24]{Proc.\@ of the 23rd International Conference
on Autonomous Agents and Multiagent Systems (AAMAS 2024)}{May 6 -- 10, 2024}
{Auckland, New Zealand}{N.~Alechina, V.~Dignum, M.~Dastani, J.S.~Sichman (eds.)}
\copyrightyear{2024}
\acmYear{2024}
\acmDOI{}
\acmPrice{}
\acmISBN{}

\acmSubmissionID{267}

\title[Forecasting and Mitigating Disruptions in Public Bus Transit Services]{Forecasting and Mitigating Disruptions in\\Public Bus Transit Services}

\author{Chaeeun Han}
\affiliation{
  \institution{Pennsylvania State University}
  \city{University Park, PA}
  \country{USA}}
\email{cfh5554@psu.edu	}

\author{Jose Paolo Talusan}
\affiliation{
  \institution{Vanderbilt University}
  \city{Nashville, TN}
  \country{USA}}
\email{jose.paolo.talusan@vanderbilt.edu}

\author{Dan Freudberg}
\affiliation{
  \institution{WeGo Public Transit}
  \city{Nashville, TN}
  \country{USA}}
\email{Dan.Freudberg@nashville.gov	}

\author{Ayan Mukhopadhyay}
\affiliation{
  \institution{Vanderbilt University}
  \city{Nashville, TN}
  \country{USA}}
\email{ayan.mukhopadhyay@vanderbilt.edu	}

\author{Abhishek Dubey}
\affiliation{
  \institution{Vanderbilt University}
  \city{Nashville, TN}
  \country{USA}}
\email{abhishek.dubey@vanderbilt.edu}

\author{Aron Laszka}
\affiliation{
  \institution{Pennsylvania State University}
  \city{University Park, PA}
  \country{USA}}
\email{alaszka@psu.edu}

\begin{abstract}
Public transportation systems often suffer from unexpected fluctuations in demand and disruptions, such as mechanical failures and medical emergencies. These fluctuations and disruptions lead to delays and overcrowding, which are detrimental to the passengers' experience and to the overall performance of the transit service. To proactively mitigate such events, many transit agencies station substitute (reserve) vehicles throughout their service areas, which they can dispatch to augment or replace vehicles on routes that suffer overcrowding or disruption. However, determining the optimal locations where substitute vehicles should be stationed is a challenging problem due to the inherent randomness of disruptions and due to the combinatorial nature of selecting locations across a city. In collaboration with the transit agency of Nashville, TN, we address this problem by introducing data-driven statistical and machine-learning models for forecasting disruptions and an effective randomized local-search algorithm for selecting locations where substitute vehicles are to be stationed. Our research demonstrates promising results in proactive disruption management, offering a practical and easily implementable solution for transit agencies to enhance the reliability of their services. Our results resonate beyond mere operational efficiency---by advancing proactive strategies, our approach fosters more resilient and accessible public transportation, contributing to equitable urban mobility and ultimately benefiting the communities that rely on public transportation the most.

\end{abstract}


\newcommand{\BibTeX}{\rm B\kern-.05em{\sc i\kern-.025em b}\kern-.08em\TeX}

\makeatletter
\gdef\@copyrightpermission{
	\begin{minipage}{0.3\columnwidth}
		\href{https://creativecommons.org/licenses/by/4.0/}{\includegraphics[width=0.90\textwidth]{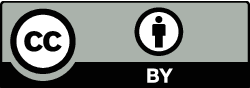}}
	\end{minipage}\hfill
	\begin{minipage}{0.7\columnwidth}
		\href{https://creativecommons.org/licenses/by/4.0/}{This work is licensed under a Creative Commons Attribution International 4.0 License.}
	\end{minipage}
	\vspace{5pt}
}
\makeatother

\begin{document}

\pagestyle{fancy}
\fancyhead{}

\maketitle

\section{Introduction}
\label{sec:intro}

\begin{figure}[b]
    \centering
    \includegraphics[width=\linewidth]{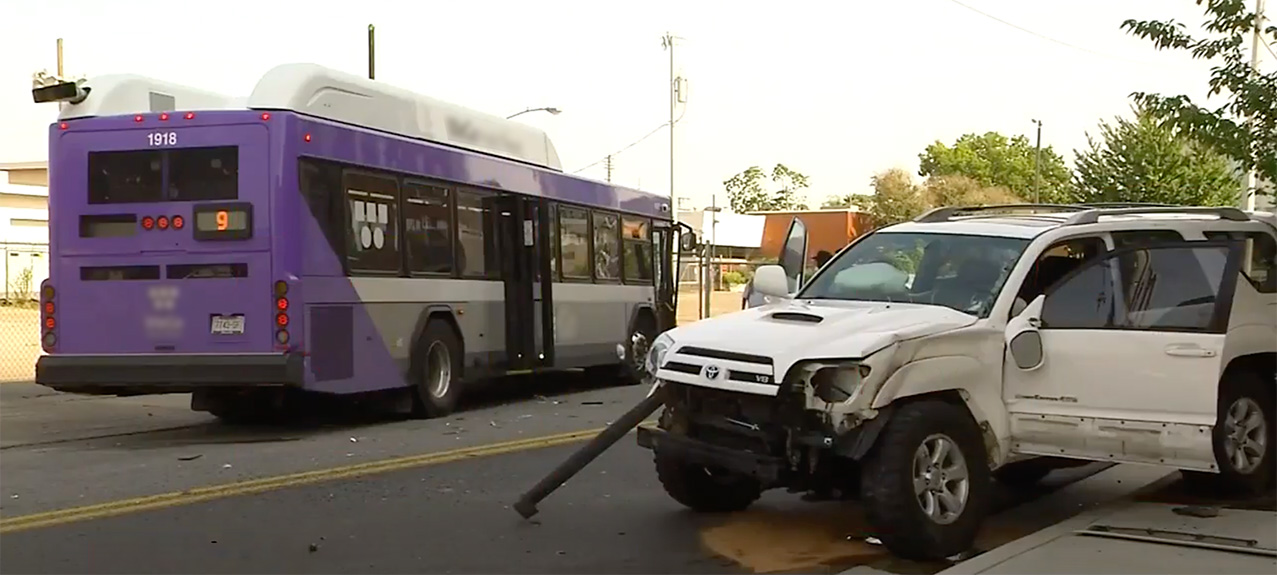}
    \caption{Disruptions, such as traffic accidents, hamper the reliability of the service provided by our partner transit agency. This image shows an actual disruption from 2020.}
    \label{fig:screenshot}
\end{figure}

Modern urban ecosystems are dependent on public transit, which gives millions of people access to mobility, influences settlements, and encourages sustainable growth. Despite their necessity, public transportation systems frequently struggle to strike a balance between available infrastructure and constantly shifting demand~\cite{ayman2022neural}. 
\Aron{added based on Abhishek's suggestion}This presents a particular challenge for mid-size U.S. cities, which typically have sparse transit networks, and where there is significant inequity in how communities rely on transit networks.
Even though it is inevitable, crowding during rush hour results in longer commute times, which is detrimental to the passengers' experience. In contrast, services tend to remain underused at non-peak times, which results in operational inefficiencies, especially in less populated suburban areas~\cite[Chapter 8.4]{rodrigue20208}.
Unexpected transit disruptions caused by a variety of reasons, such as vehicle breakdowns, operational issues, or medical emergencies, compound these difficulties (\cref{fig:screenshot}). 
These disruptions cause cascade effects across the entire transit network by extending travel times and disturbing established traffic patterns~\cite{sun2016estimating}. Both riders and operators are heavily impacted by these disruptions, highlighting the urgent need for proactive mitigation methods.\ad{should we provide some metrics on the importance of efficient public transportation on the community}

The problem setting that we consider has two distinct challenges: first, we seek to forecast disruptions in space and time, which we refer to as the \textit{forecasting problem}, and second, we seek to optimize the stationing of substitute buses to ensure that they can promptly respond to disruptions, which we refer to as the \textit{stationing problem}. 
However, forecasting disruptions is challenging due to the scarcity of data. The relative rarity of disruptions means there is a limited amount of data to draw conclusions from using conventional machine learning models~\cite{yap2021predicting}. While prior work has explored the impact of disruptions on public transportation~\cite{yap2020measuring, wang2023modeling, ghaemi2018impact}, there is significantly less work on attempting to predict disruptions~\cite{yap2019analysis}. 
While \citet{yap2019analysis} recently explored disruptions in public transit, our setting is significantly more challenging from the perspective of data availability---\citet{yap2019analysis} focus on disruptions in metro (subway) networks, which have more than 100 types of disruptions, as opposed to only a handful in the case of bus networks (our partner agency has 20 types of disruptions), and the frequency of disruptions in metro networks is almost thrice as high as in our problem setting.

As for the challenge of bus stationing,
based on conversations with our partner transit agency, we learned that there are no formal policies for where buses are to be stationed at the start of each day. 
Agencies rely on a combination of historical ridership data and domain expertise to identify potential pain points across the city. 
Currently, our partner agency stations its buses at locations mainly near the central terminal and at high-frequency lines which experience consistent overcrowding.
The stationing problem is challenging due to the combinatorial nature of the search space (i.e., substitute vehicles can be stationed at arbitrary stops) and the inherent uncertainty of the disruptions. The problem therefore must maximize expected utility\footnote{We define utility later, while formalizing the problem.} while considering a multitude of factors such as passenger demand, geographic locations, traffic patterns, and operational constraints~\cite{rajesh2022multi}\ad{important to clarify if there is some factor that we are not using but is being mentioned here}. As the combinatorial search space grows exponentially with the growth of urban areas, transit agencies need scalable approaches to dynamically station vehicles. Most prior work has looked at this problem from a purely demand-responsive setting~\cite{zhou2023optimizing} (where passengers reserve rides in buses, similar to taxis), which is fundamentally different from stationing buses for mitigating the effects of unexpected disruptions.

In this paper, we tackle the twin challenges of forecasting disruptions  and stationing substitute vehicles  in collaboration with the transit agency of Nashville, TN, a mid-size U.S. city. Specifically, we seek to estimate the likelihood of disruptions in fixed-line bus transit, and given the likelihood, optimize the stationing of substitute buses to mitigate the effects of disruptions. By using our approach, the transit agency will identify routes that are prone to disruptions and ensure that the optimal allocation of substitute buses will minimize the effects on passengers and increase operational efficiency. Specifically, we make the following contributions: \textbf{1)} We create data-driven statistical models that can estimate the probability that a given future transit trip will encounter a disruption. \textbf{2)} We formulate the stationing problem and introduce efficient heuristic algorithms to find stationing locations that minimize the impact of disruptions on passenger experience.

The remainder of this paper is organized as follows.
\cref{sec:model} formulates the problem of forecasting disruptions and optimizing vehicle stationing in public transit.
\cref{sec:approach} introduces our proposed computational approaches for forecasting disruptions and optimizing stationing.
\cref{sec:experiment} describes our real-world data and experimental setup.
\cref{sec:results} presents numerical results, demonstrating the accuracy of our forecasts and the improvement in transit-service operations due to optimizing stationing.
\cref{sec:related} gives a brief overview of related work.
Finally, \cref{sec:concl} provides concluding remarks.

\section{Problem Formulation}
\label{sec:model}

In this section, we describe the characteristics of fixed-line transit, introduce disruption forecasting, and formulate the vehicle stationing problem. 
Fixed-line public transit relies on a set of vehicles running on a consistent and predictable schedule of trips to service passengers. Any events or incidents that delay or disrupt buses can impact their reliability leading to increased wait times, interfere with transfers, and increase uncertainty. As a contingency, transit agencies have a limited number of substitute buses, which are deployed in response to disruption and overcrowding events.
Having the ability to forecast possible disruptions allows transit agencies to station substitute buses in anticipation of such events.

\textbf{Trip:} Fixed-line transit vehicles follow a scheduled assignment of trips. A trip pertains to individual occurrences of a bus navigating its assigned route. For a given route $r \in R$, there is a set $T(r)$ containing all trips on the route. A trip $t$ represents a specific instance of a bus traveling a route in a particular direction at a given time. For example, the trip $t$ of the first bus operating on Sundays for route~$r$ symbolizes a specific occurrence, distinct from other trips on that~route.

\textbf{Route + Direction:} A bus route, combined with its direction, is a group of trips that represents a predetermined pathway recurrently followed. The set $R$ contains all unique combinations of routes and directions in the system, where each combination $r \in R$ uniquely signifies a specific pathway and direction. For example, outbound and inbound buses might share identical stops, but since they have different points of origin and termination, they are distinct in~$R$.

\textbf{Stop}: Bus stops are strategically positioned along the route, and they enable the bus service to cater to a wide spatial region.
For a given trip $t \in T(r)$ on route $r \in R$, there is a set $S(t)$ of stops for that trip.
For a mid-size U.S. city, it is common for some routes to share one or more stops. This sharing often aligns with many transit agencies' practice of following a spoke-hub model which is often cheaper to develop~\cite{DAGANZO2010434}, where most routes have at least their origin or destination in a central, downtown terminal.

\subsection{Disruption Forecasting}
\label{sec:model_forecasting}

To assess the likelihood of a disruption occurring, we formulate disruption forecasting as a binary classification problem.
However, rather than merely categorizing trips as having a disruption (1) or not (0), we focus on estimating the probability that a given trip will be classified as having a disruption. This approach enables a more granular understanding of the risks associated with different trips. 

We assume that we have a dataset of past trips represented as $T = \{t_1, t_2, ..., t_n\}$. Each trip $t_i \in T$ is characterized by a vector of features $X_{t_i}$, which includes categorical and numerical variables.
Categorical variables include route identifier combined with its direction($R_{t_i}$) which is a specific route number followed by its direction name, %
ridership ($P_{t_i}$), passenger occupancy on the trip, %
service-time windows ($S_{t_i}$),
and calendar-related variables representing the date of the trip that contains year ($Y_{t_i}$), month ($M_{t_i}$), and day of week ($W_{t_i}$).
Numerical variables capture weather information such as precipitation intensity ($I_{t_i}$) and temperature. ($T_{t_i}$)
Thus, we can represent $X_{t_i}$ as follows:
\begin{equation}
    X_{t_i} = (R_{t_i}, P_{t_i}, S_{t_i}, Y_{t_i}, M_{t_i}, W_{t_i}, I_{t_i}, T_{t_i})
\label{eq:vector}
\end{equation}
Also, each $t_i$ has a corresponding label $y_{t_i}$ indicating whether a disruption was reported during the trip.

The goal of the prediction model is to learn $\Pr[y_{t_i}=1 \mid X_{t_i}]$ from a function $f$ mapping the feature vectors to the labels, $f: X_{t_i} \rightarrow y_{t_i}$, using the dataset as the training data.
However, the inherent randomness of disruptions, coupled with the fact that the occurrences of disruptions are very rare compared to the number of bus operations, inevitably makes forecasting disruptions a challenging~task.\ad{we should probably clarify that due to sparsity we cannot do a joint distribution on trip and stop}

\subsection{Vehicle Stationing}
\label{sec:model_stationing}

On a typical transit day, the transit agency has access to two different types of buses, regular and substitute buses. \textbf{Regular buses} are buses that have been scheduled to serve trips across the city for a given day. Once a regular bus has finished its shift for the day, they are not available for dispatch. Regular buses travel across their designated route, picking up and dropping off passengers at stops along their trip. Collectively, buses will pick up and drop passengers across $J$ stops in the city. Passengers can be affected due to two types of events: first, at times, buses cannot accommodate more passengers due to crowding, resulting in \textit{overage events} that are temporary, or second, buses can encounter disruptions such as mechanical failures or accidents that render them unusable for the rest of the day, which we refer to as \textit{disruption events}. \textbf{Substitute buses} are buses that are held in reserve for the purpose of being dispatched to cover either of these events. These start their day at the main depot, travel to a predetermined stationing stop to wait for dispatching, and then end the day at the main depot. Substitute buses traveling from the main depot to a stationing stop, or from a stationing stop to a dispatch location, are not considered to be in service. The distance traveled and time spent by buses when not in service are called deadhead miles and deadhead time respectively.

Our conversations with the transit agency reveal that existing stationing strategies are largely \textit{ad-hoc}, i.e., there are no formal and established policies for stationing. Buses are traditionally stationed by leveraging summary statistics from historical data (e.g., frequency of disruptions on a route) and domain expertise of transit operators. Similarly, there are no principled policies for dispatching substitute buses; buses are dispatched simply based on availability, with higher priority given to disruption events over overage events. 
Due to this, experts often consider contexts such as headways, traffic congestion, and forecasted ridership, before they either \textbf{dispatch a substitute vehicle} or \textbf{do nothing} in response to an~event. 

We present a principled optimization formulation for the stationing problem. Formally, consider a subset of stops $S_\textit{station}$ where buses can be stationed using the agency's budget of $k$ substitute buses. A feasible solution $\bar{x}$ is a $k$-subset of $S_\textit{station}$ (i.e., $\bar{x} \subseteq S_\textit{station}$ such that $|{\bar{x}}| = k$).
Our goal is to find the subset of stops to station the substitute buses that will minimize their total non-service miles ($D$) and non-service travel duration ($T$) while minimizing the total passengers left behind ($L$) across all stops. Formally, the problem is:
\begin{equation}
\operatorname{argmin}_{\bar{x} \subseteq S_\textit{station} \,:\, |\bar{x}|=k} \mathbb{E}_P \left [ D(\bar{x}; P) + T(\bar{x}; P) + \sum_{j=1}^{J} L(j, \bar{x}; P) \right]
\label{eq:objective}
\end{equation}
where $P$ is the random variable that denotes the occurrence of disruptions and overcrowding events. It is worth noting that this objective is practically impossible to calculate based on first principles. Thus, in practice, we approximate its value by simulating a number of outcomes.

\section{Proposed Approach}
\label{sec:approach}

Our approach consists of two components: \textbf{1)} We employ a multifaceted approach to predict the likelihood of trip disruptions under given conditions, utilizing a combination of statistical and machine learning methodologies, including logistic regression~\cite{bishop2006pattern}, extreme gradient boosting (XGBoost)~\cite{xgboost}, and Poisson regression~\cite{hilbe2014modeling}.
\textbf{2)} We identify the best subset of stops for stationing through a combination of greedy search and simulated annealing~\cite{vanLaarhoven1987}.

\subsection{Disruption Forecasting}
\label{sec:approach_forecasting}

Forecasting the likelihood of disruption occurring on a certain trip is a supervised learning problem. 
We explore two learning-based approaches, namely logistic regression and the XGBoost model. 
Using the vector of features described in Equation (\ref{eq:vector}), the logistic regression model can be expressed as:
\iAron{We need to be careful with our usage of symbols: JP used $\beta$ (and $\alpha$, which will be a problem for Poisson regression) to denote something different.}
\begin{equation}
\Pr\left[y_{t_i} = 1 | X_{t_i}\right] = \frac{1}{1 + e^{-\beta_0 - \beta_1 \cdot X_{t_i}}}
\end{equation}
where $y_{t_i}$ is the binary outcome (presence of disruption) for trip $t_i$, and $\beta_0$ and $\beta_1$ are the parameters to be estimated.
Note that our problem focuses on estimating the likelihood of disruptions in a trip and does not address stop-level prediction; naturally, predicting which stop might face a disruption is practically infeasible due to the sparsity of data. \ad{yes in my comment on problem definition we should add this statement there as well}
The logistic regression is advantageous in that it is simple and interpretable, so it is well suited for understanding the impact of individual features on the outcome.

Also, the XGBoost model is employed with the same problem settings as logistic regression. 
It is employed for its robust performance and ability to handle complex nonlinear relationships. Its gradient boosting framework enables the assembly of an ensemble of weak predictive models, making it capable of capturing intricate patterns in the data~\cite{chen2016xgboost}.
Moreover, XGBoost provides a feature importance score, offering valuable insights into factors significantly influencing disruption likelihood. 
However, the raw output probabilities may not always represent the true likelihood of events. Calibration techniques, such as Isotonic Regression, can be employed post hoc on the model to adjust the probabilities to better reflect the true outcomes, thereby increasing their reliability in practical scenarios. It is a non-parametric approach that can fix both over-confidence and under-confidence cases well~\cite{song2021classifier}.

\subsection{Vehicle Stationing}
\label{sec:approach_stationing}

\iAron{try GRASP-style approach? \url{https://link.springer.com/referenceworkentry/10.1007/978-3-319-07153-4_23-1}}

\begin{algorithm}[!t]
\small
\caption{Greedy Selection}
\label{alg:greedy}
\begin{algorithmic}[1]
\State $C \leftarrow$ List of candidate stops
\State $C^* \leftarrow \{\}$
\State $k \leftarrow$ Number of substitute buses
\While{$count(\bar{C}) < k$}
    \State $C \gets C \setminus \bar{C}$
    \State $\hat{c} \leftarrow \{\}$
    \State $O = \infty$
    \For{$c \in C$}
        \State $\hat{C} \leftarrow c$
        \State $\hat{O} = cost(\hat{C})$  \Comment{objective function}
        \If{$\hat{O} < O$}
            \State $\hat{c} = c$
            \State $O = \hat{O}$
        \EndIf
    \EndFor
    \State $C^* \leftarrow \hat{c}$
\EndWhile
\State \Return $\bar{C}$
\end{algorithmic}
\end{algorithm}

\begin{algorithm}[!t]
\small
\caption{Neighbor Generation}
\label{alg:neighbor}
\begin{algorithmic}[1]
\State $C \leftarrow$ Candidate stops
\State $\bar{x} \leftarrow$ Current solution
    \State $i = random(0, count(S))$ \Comment{uniformly at random}
    \State $\bar{x}[i] = random(C \setminus \bar{x})$ \Comment{uniformly at random}
\State \Return $\bar{x}$
\end{algorithmic}
\end{algorithm}

Selecting the best locations to station substitute buses is a combinatorial optimization problem that can not be reasonably solved in linear time due to the exponential possibilities available (over 1000 possible stops across the city). Instead, we select a subset of 25 stops for stationing based on the current and potential stationing locations that the transit agency uses.
We then generate an initial stationing plan using a greedy algorithm, \cref{alg:greedy}, that selects the best solutions using the same objective function in \cref{eq:objective}. The resulting stationing plan is then used as the initial solution to simulated annealing.
In simulated annealing, neighboring state iterations are then generated 
by randomly selecting a single bus, and then assigning it to a new random stop, \cref{alg:neighbor}, with the constraint that no two buses should be at the same stop at the same time. \cref{alg:simulated_annealing} details how we select the solution set that gives the minimum cost after $N$ iterations, where mutations are dependent on the initial temperature selected and the current iteration.

\begin{algorithm}[!t]
\small
\caption{Simulated Annealing Optimizer}
\label{alg:simulated_annealing}
\begin{algorithmic}[1]
\State $N \leftarrow$ Simulated annealing iterations
\State $K \leftarrow K_{max}$
\State $\bar{x} \leftarrow greedy(x)$ \Comment{select initial solution using \cref{alg:greedy}}
\State $O \leftarrow cost(\bar{x})$ \Comment{objective function}
\For{$n=0$ through $N$}
    \State $\bar{x}_{new} = neighbor(\bar{x})$
    \State $O_{new} = cost(S^*_{new})$
    \State $\Delta{O} = O_{new} - O$
    \If{$accept(\Delta{O}, K)$}
        \State $\bar{x} \leftarrow \bar{x}_{new}$
        \State $O \leftarrow O_{new}$
    \EndIf
    \State $K = \frac{K}{\gamma + n}$ \Comment{cool the temperature}
\EndFor
\State \Return $\bar{x}$ \Comment{best stationing plan}
\end{algorithmic}
\end{algorithm}

\section{Experimental Setup}
\label{sec:experiment}

\subsection{Real-World Data}
\label{sec:data}

To conduct our analysis of disruption prediction, we used disruption data from the transit agency of Nashville, TN,  
transit schedule data in the form of General Transit Feed Specification (GTFS), Automated Passenger Count (APC) data, and weather data. 

\textbf{Automated Passenger Count data}
APC data records a variety of information that transit agencies use to monitor bus conditions. It uses infrared sensors along doors which trigger whenever they open for passengers. It records the number of people boarding and alighting at every bus stop along its trip. It also records the actual times it arrives and departs at every stop. This allows operators to react to bus bunching events by adding slack at certain points on the trip or by dispatching substitute buses.

\textbf{Disruption data}
The disruption dataset contains comprehensive information about each disruption, including where and when it occurred and the underlying cause. This information includes locations specified by stop names as well as latitude and longitude coordinates, corresponding to the last stop where the bus terminates operation before the disruption.
These locations facilitated the integration of disruption data with APC and weather information, ensuring consistency across datasets.
This dataset covers the period of March 2020 to January 2023, during which time a total of 5,096 disruptions were recorded.

\textbf{GTFS data}
The General Transit Feed Specification (GTFS) dataset offers insights into bus schedules, encompassing all planned transit trips. By aligning this data with disruption data, we could investigate the frequency and patterns of disruptions.

Each transit trip is characterized by unique identifiers depending on the day of the week, time of day, and bus route. This allows us to analyze disruptions with a trip of specificity. As there are no trips between midnight and 4AM, we segmented the 20 service hours into five time windows:
early morning (4AM to 6AM), morning (6AM to 9AM), mid-day (9AM to 2PM), afternoon (2PM to 6PM), and evening (6PM to midnight). 
For the ridership, we divided occupancy rate into four categories: low (0\% to 30\%), moderate (30\% to 60\%), high (60\% to 100\%), and over-capacity ($>100\%$). 

\begin{table}[]
\footnotesize
\centering
\caption{Overview of Datasets}
\begin{tabular}{|l|l|l|l|}
\hline
\textbf{Data} & \textbf{Source} & \textbf{Scope} & \textbf{Features}     \\ 
\hline
\hline
APC & Agency & 2020-2023 & {ridership, stops} \\
\hline
Disruption & Agency & 2020-2023 & {GPS location, datetime, stops}\\ \hline
GTFS~\cite{mchugh_GTFS}  & Agency  & 2020-2023 &  {routes, stops}          \\ \hline
Weather & Darksky & 2020-2023 & {location, temperature, precipitation} \\ \hline
\end{tabular}
\label{table:dataset_overview}
\end{table}

\subsection{Disruption Forecasting}
The disruption forecasting model uses all data shown in Table~\ref{table:dataset_overview}. For six possible categorical variables which are route identification combined with direction, service-window hours, day of week, passenger capacity, year, and month, we use one-hot encoding to train in the models. 
For the logistic regression model, we conduct the goodness-of-fit test and build 64 ($2^6$) models using all possible combinations of the variables. The goodness-of-fit test of the logistic regression model offers insight into how various independent factors affect the probability of disruptions. So, we are able to choose the optimal combination among categorical variables.
After deciding what variables are to be used for the model, we compare the performance of three disruption forecasting models; logistic regression, XGBoost, and XGBoost with Isotonic Calibration. 
For the experiment, we used Python 3.9.12 and Scikit-learn 1.3.0~\cite{scikit-learn} for training.

\subsection{Simulator}
\label{sec:simulator}

The simulator uses data from APC and GTFS to recreate public transit activities for a single day. There are three main components to the simulator: regular buses, substitute buses, and passengers. 
Regular buses start and end their day at the main depot after servicing all their trips. They may pick up or drop off passengers at every stop along their trip. 
Substitute buses start their day at the main depot, travel to their predetermined stationing location, and then end at the main depot. When dispatched to cover for a regular bus, they must travel from their current location to the stop in question, accumulating deadhead miles for a period of time. When covering an overage event, the substitute bus will travel from its current location to some stop $\hat{s}$ where the event was reported. From then on, it will travel the same trip as the bus that reported the overage, until the final stop in the trip. Only upon finishing this would it be available for dispatch again. When dispatched to a bus that has broken down, it travels to the last visited stop of the broken bus, and from then on, covers all the trips that the broken down bus was meant to service. Only upon finishing this sequence of trips would the substitute bus be available for dispatch.
Substitute buses are dispatched following a policy. The bus that is \textit{nearest to a reported event will be considered for dispatch}, and ties are broken randomly. If the event that triggered this policy is an \textit{overage event}, then the policy will only dispatch the bus if the number of people left behind is greater than 5\% of the current bus' capacity. However, disruption events are handled immediately, if there are available substitute buses. Every time a substitute bus is dispatched, it accrues deadhead miles and deadhead time. 

At each stop, a certain number of \textbf{passengers} can either board or alight the bus. Passengers can arrive at the stop as early as 10 minutes before the bus arrives and will wait for the bus for a maximum of 30 minutes. Passengers can board any bus arriving at their current location as long as the bus is traveling in the same block, route, and direction, and if bus ridership is less than bus capacity. If passengers are unable to board the bus, they remain at that stop. If a disruption event occurs, passengers currently onboard the bus are unloaded at the last passed stop and will wait for another 30 minutes for another bus to arrive. Any trip on the service day can encounter a disruption, causing the bus to fail at one of the stops along the trip, ending the service. A day can have anywhere from zero disruptions to a disruption at each of its trips. 

We address stochasticity in the real world by generating chains of events for each single day.   For each chain, we sample passenger counts and disruptions independently. Generative models, trained over ridership at the stop level~\cite{talusan2022apc}, are used to sample the ridership forecast at each stop. Boarding and alighting counts are then derived from this value. Disruption events are sampled from the disruption forecast model discussed in \cref{sec:model_forecasting}. We then get the average of performance of our stationing across all the chains for each day.

This is an event-driven simulator that considers bus arrivals at stops, overages, and disruptions as events that require a dispatch decision. The simulator begins with the first passengers arriving at a stop and proceeds until the final bus returns to the depot for the last stop of the trip. Passengers who have not been picked up at the end of the simulation time are considered to have been left behind. For this paper, we truncate the service day to all trips that start between 6AM 
and 1PM. Table~\ref{table:simulation_params} lists the rest of the configurations used in the experiment. The initial temperature was selected after a hyperparameter search.
All simulation experiments were done on Chameleon servers~\cite{keahey2020lessons} with 187 GB of RAM and an Intel Xeon CPU with 96 cores, each with 2.4GHz.

\begin{table}[]
 \small
\centering
\caption{Simulation Parameters}
\begin{tabular}{|l|c|}
\hline
\textbf{Parameter}              & \textbf{Setting}      \\ \hline
\hline
simulation duration             & 6AM to 1PM  \\ \hline
passenger arrival range         & within 10 minutes     \\ \hline
passenger patience              & 30 minutes            \\ \hline
overage threshold               & $\geq105\%$           \\ \hline
average number of buses per day & 80.0                  \\ \hline
number of substitute buses      & 5                     \\ \hline
number of chains                & 100                   \\ \hline
number of days                  & 50                    \\ \hline
initial temperature             & 100                   \\ \hline
local search iterations         & 500                   \\ \hline
\end{tabular}
\label{table:simulation_params}
\end{table}

\section{Numerical Results}
\label{sec:results}

\subsection{Disruption Forecasting}
\label{sec:results_forecasting}

\begin{table*}[htbp]
\centering
\caption{Result of logistic regression models with different feature sets variables}
\label{tab:nll_models}
\def\arraystretch{1.1}
\small
\begin{tabular}{|l|c|c|}
\hline
\textbf{Features (Number of features)} & \textbf{Train Cross Entropy} & \textbf{Test Cross Entropy} \\ 
\hline
\hline
Route + Direction, Service Window Hours, Day of Week, Passenger Capacity, Year, Month (6) &  0.1320 & 0.1330\\ 
\hline
Route + Direction, Service Window Hours, Day of Week, Passenger Capacity, Month (5) &  0.0903 & 0.0912 \\ 
\hline
Route + Direction, Service Window Hours, Day of Week, Passenger Capacity (4) & 0.0905  & 0.0913\\ 
\hline
Route + Direction, Service Window Hours, Passenger Capacity (3) & 0.0946 &0.0952 \\
\hline
Route + Direction, Passenger Capacity (2) & 0.0953 & 0.0960\\ 
\hline
Route + Direction, Service Window Hours, Day of Week, Passenger Capacity, Month, Temperature (6) & 0.0903 & 0.0910 \\ 
\hline
\end{tabular}
\end{table*}

\paragraph{Model Performance}
The performance of the predictive models is evaluated using cross entropy as a key metric, where lower values signify better model performance. 
We compare cross-entropy over accuracy due to the dataset’s imbalance, revealing that just 0.2\% of all incidents are disruptions ($y_{t_i} = 1$). Accuracy, skewed by the abundance of non-disruptions ($y_{t_i}=0$), could mislead. Therefore, cross-entropy, being a more informative metric, is used to accurately gauge the model's classification performance, especially concerning the minority class.

As indicated in Table~\ref{tab:cross_entropy}, three models undergo evaluation: Logistic Regression, XGBoost, and XGBoost calibrated with Isotonic Regression. XGBoost classifier calibrated with Isotonic regression shows the best performance, attributed to Isotonic Calibration’s refinement of the XGBoost probability estimates, aligning predictions more closely with actual outcomes.
A grid search for the best hyperparameter values for the XGBoost model found the model performs the best when the learning rate is 0.1, maximum depth is 9, minimum child weight is 1, and subsample is 0.5.

\begin{table}[t]
\footnotesize
\centering
\caption{Results for three disruption forecasting models}
\begin{tabular}{|l|l|l|} \hline
\textbf{Model} &  \textbf{Train Cross Entropy} & \textbf{Test Cross  Entropy}\\ \hline \hline
Logistic Regression &  0.0903& 0.0910\\ \hline 
XGBoost&  0.0872& 0.0881\\ \hline 
XGBoost + Isotonic Calibration &  0.0870& 0.0876\\ \hline
\end{tabular}
\label{tab:cross_entropy}
\end{table}

\paragraph{Model and Feature Selection}

Five categorical features that are route identifier combined with direction, service-window hours, month, and passenger capacity, combined with numerical features which are weather related-variables like temperature and precipitation shows the optimal result. In detail, logistic regression model – one with five categorical features, excluding only the year, demonstrates the lowest cross entropy value as shown in \cref{tab:nll_models}.
Subsequently, an XGBoost prediction model outperformed our logistic regression model, offering a more refined prediction with an even lower cross-entropy value.

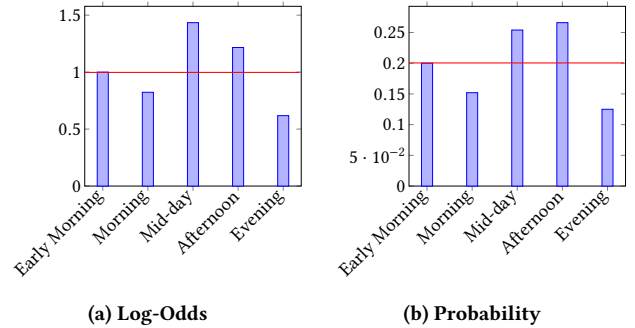
\begin{figure}[ht]
\pgfplotstableread[col sep=comma]{
Feature,Log-Odds,Prob
Early Morning,1.000,0.20
Morning,0.823,0.152
Mid-day,1.434,0.254
Afternoon,1.216,0.266
Evening,0.618,0.125
}\datatable
\centering
\begin{subfigure}{0.44\columnwidth}
\centering
\begin{tikzpicture}[scale=0.42]
\begin{axis}[
    ybar,
    bar width=0.35cm,
    symbolic x coords={Early Morning,Morning,Mid-day,Afternoon,Evening},
    xtick=data,
    x tick label style={rotate=45, anchor=east},
    ymin=0,
    ticklabel style={font=\Huge}
]
\addplot table [x=Feature, y=Log-Odds]{\datatable};
\end{axis}
\draw[red] (0, 3.6) -- (6.8, 3.6);
\end{tikzpicture}
\caption{Log-Odds}
\label{fig:log-odds}
\end{subfigure}
\hspace{0.05\columnwidth}
\begin{subfigure}{0.44\columnwidth}
\centering
\begin{tikzpicture}[scale=0.42]
\begin{axis}[
    ybar,
    bar width=0.35cm,
    symbolic x coords={Early Morning,Morning,Mid-day,Afternoon,Evening},
    xtick=data,
    x tick label style={rotate=45, anchor=east},
    ymin=0,
    ticklabel style={font=\Huge}
]
\addplot table [x=Feature, y=Prob]{\datatable};
\end{axis}
\draw[red] (0, 3.9) -- (6.8, 3.9);
\end{tikzpicture}
\caption{Probability}
\label{fig:probability}
\end{subfigure}
\caption{Log-Odds and Probability Values for Features.}
\label{fig:combined}
\end{figure}

Moreover, notably, midday to afternoon service-window hours were more susceptible to disruptions compared to late or early hours as illustrated in \cref{fig:combined}. Evening hours shows the lowest probability of disruptions. We analyze this impact of each feature by examining the log-odds ratio for route, service-window hours, and month. Positive odds ratio indicates a higher likelihood of disruptions.

Lastly, permutation tests are conducted to validate the significance of these findings. As shown in Table~\ref{tab:permutation_test}, the result indicated that disruptions frequencies among routes follow the same distribution, affirming that our analysis is not accounting for potential data imbalances. These outcomes underscore the robustness of our predictive model, offering valuable insights into understanding and forecasting disruptions in real world scenarios.

\begin{table}[t]
\footnotesize
    \caption{$p$-value of Permutation Test for Number of Disruptions on Trips per Route}
    \centering
    \begin{tabular}{|c||c|c|c|c|c|c|c|} \hline 
         \textbf{Route}& \textbf{3}&  \textbf{4}&  \textbf{5}&  \textbf{6}&  \textbf{7}&  \textbf{8} &\textbf{9}\\ \hline \hline
         3&  1.0&  -&  -&  -&  -&   -&-\\ \hline 
         4&  0.4818&  1.0&  -&  -&  -&   -&-\\ \hline 
         5&  0.6501&  0.7945&  1.0&  -&  -&   -&-\\ \hline 
         6&  0.7744&  0.8496&  1.0&  1.0&  -&   -&-\\ \hline 
         7&  0.6393&  0.1294&  0.1793&  0.2252&  1.0&   -&-\\ \hline 
         8&  1.0&  0.4005&  0.5305&  0.5778&  0.5489&  1.0 &-\\ \hline 
         9&  0.6683&  0.3148&  0.2746&  0.3366&  0.8008&  0.5264 &1.0\\\hline
    \end{tabular}
    \label{tab:permutation_test}
\end{table}

\begin{figure*}[ht!]
\centering
\begin{tikzpicture}
\begin{axis}[
    legend style = {at={(0.5, -0.3)}, anchor=north, legend columns=0},
    ybar,
    x tick label style={rotate=45, anchor=east, align=center},
    width=\linewidth,
    height=6cm,
    bar width=0.2cm,
    bar shift = 0pt,
    symbolic x coords={
        Early Morning, Morning, Mid-day, Afternoon, Evening,
        High, Low, Moderate, Overload,
        January, February, March, April, May, June, July, August, September, October, November, December,
        Route 1, Route 2, Route 3, Route 4, Route 5, Route 6,
        Monday, Tuesday, Wednesday, Thursday, Friday, Saturday, Sunday
    },
    xtick={
        Early Morning, Morning, Mid-day, Afternoon, Evening,
        High, Low, Moderate, Overload,
        January, February, March, April, May, June, July, August, September, October, November, December,
        Route 1, Route 2, Route 3, Route 4, Route 5, Route 6,
        Monday, Tuesday, Wednesday, Thursday, Friday, Saturday, Sunday
    },
    xticklabel style={font=\normalsize, text width=2cm, align=right},
    ymajorgrids=true,
    enlarge x limits=0.02,
    ymin=0,
]

\addplot+[ybar] plot coordinates {
    (Early Morning, 179)
    (Morning, 349)
    (Mid-day, 520)
    (Afternoon, 484)
    (Evening, 229) 
    };
\addplot+[ybar] plot coordinates {
    (High, 221)
    (Low, 562)
    (Moderate, 356)
    (Overload, 97) 
    };
\addplot+[ybar] plot coordinates {   
    (January, 185)
    (February, 233)
    (March, 325)
    (April, 189)
    (May, 170)
    (June, 199)
    (July, 215)
    (August, 233)
    (September, 205)
    (October, 247)
    (November, 206)
    (December, 212) 
    };
\addplot+[ybar] plot coordinates {
    (Route 1, 110)
    (Route 2, 100)
    (Route 3, 104)
    (Route 4, 93)
    (Route 5, 102)
    (Route 6, 98) 
    };
\addplot+[ybar] plot coordinates {    
    (Monday, 383)
    (Tuesday, 425)
    (Wednesday, 371)
    (Thursday, 374)
    (Friday, 424)
    (Saturday, 202)
    (Sunday, 193)
};
\legend{Service Window Hours, Passenger Capacity, Month, Route + Direction, Day of Week}

\end{axis}
\end{tikzpicture}
\caption{Feature importance of one-hot encoded categorical features used for XGBoost model.}
\label{fig:feature_importance}
\end{figure*}
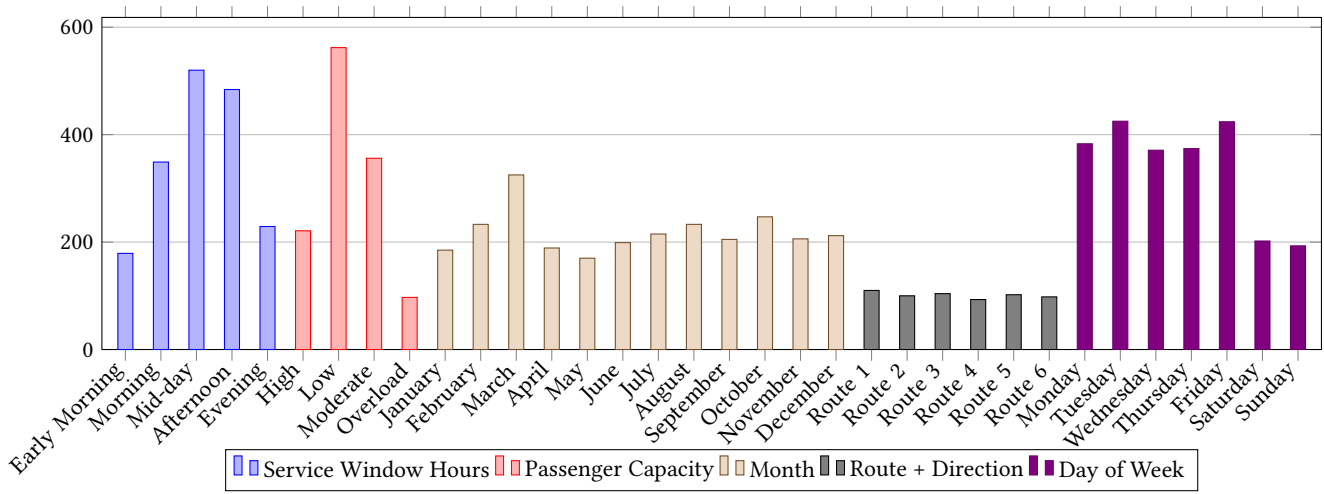

\paragraph{Features Importance}
The XGBoost model's feature importance scores, presented in \cref{fig:feature_importance}, identified service-window hours as the most crucial predictor of disruption likelihood. 
According to the feature-importance results, service-window hours were determined to be the most important feature in this instance, followed by day of week, month, and route ID. These observations can help direct future research and inform decisions based on the output of the~model.

\subsection{Vehicle Stationing}
\label{sec:results_stationing}

We compare our results to the three different stationing assignments that the public transit agency uses. \textbf{Garage} represents the absence of a stationing plan, all substitute buses are kept at the main garage. \textbf{Hub} represents a stationing plan where all substitute buses are stationed at the central hub. Finally, \textbf{Agency}, represents the current stationing plan being used by the transit agency, based on our conversations with them. 

\begin{figure}[t]  
    \centering
    \includegraphics[width=0.75\linewidth]{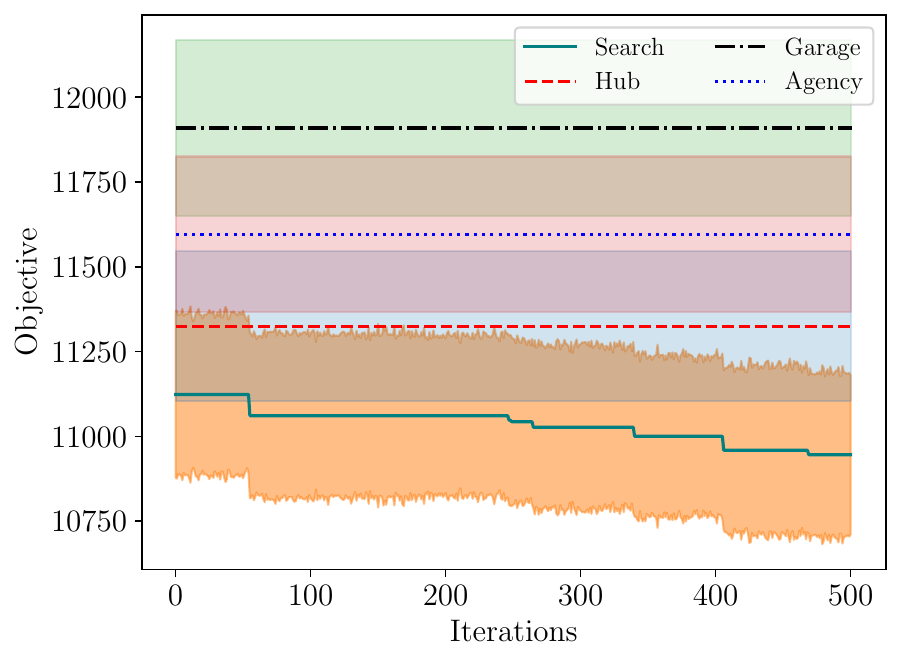}
    \caption{Comparison of cost using stationing plans based on search, garage, or agency. Lines and shaded regions represent mean and standard error across 100 chains.}
    \label{fig:iterations}
\end{figure}

\begin{figure}[t]  
    \centering
    \includegraphics[width=0.75\linewidth]{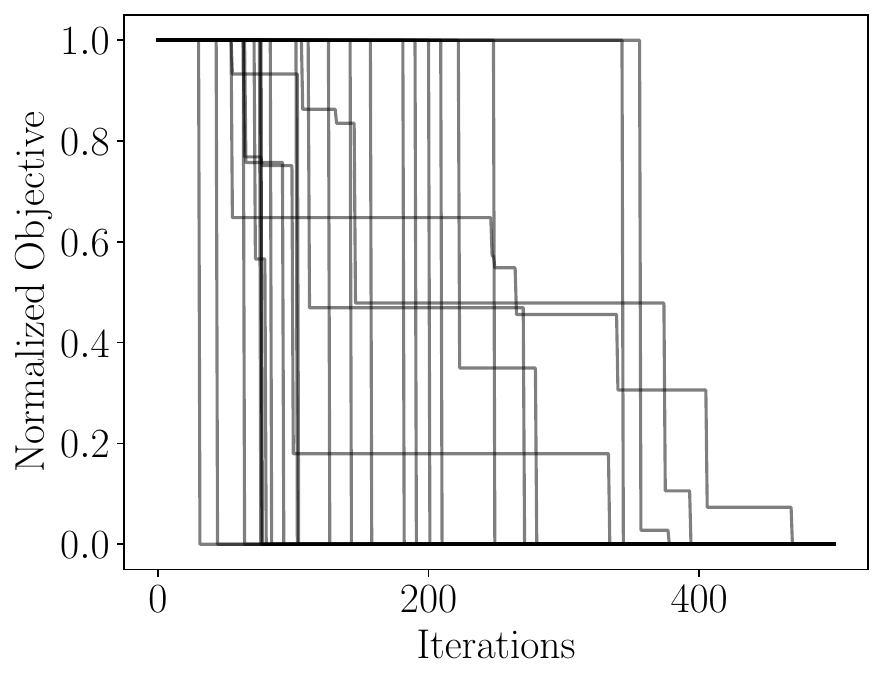}
    \caption{Normalized objective costs for all test days.}
    \label{fig:all_days}
\end{figure}

\begin{figure}[t]  
    \centering
    \includegraphics[width=0.75\linewidth]{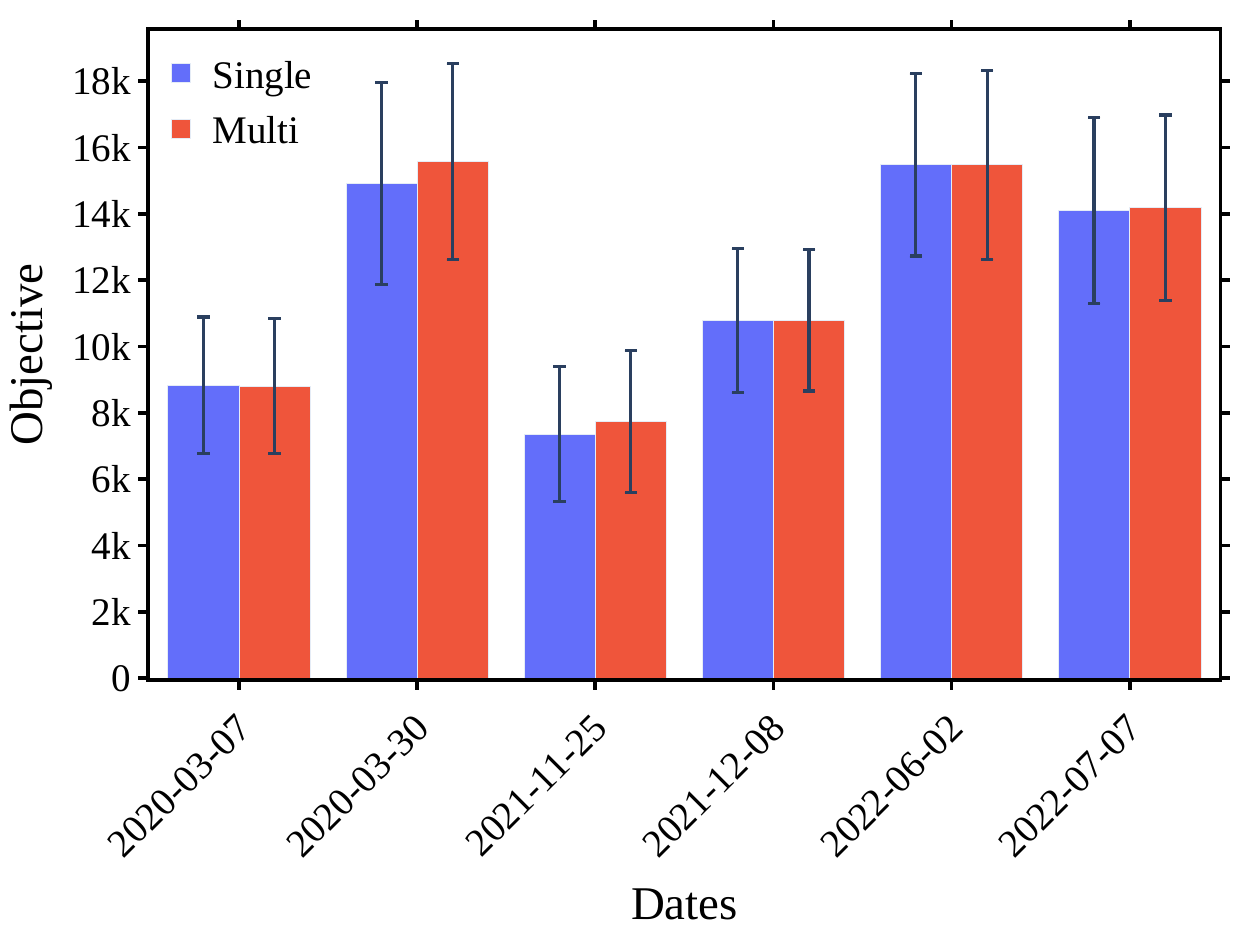}
    \caption{Comparison of daily optimization for six days and optimizing for all six days at once. Lower is better.}
    \label{fig:multi_day}
\end{figure}

\paragraph{Simulation and random local search} For every iteration of the random local search, we simulate the same day with 100 different chains of events. Figure~\ref{fig:iterations} shows the progress of the objective scores for a single day. The initial allocation generated by the greedy approach results in a lower cost than the baselines, and within 50 iterations, we can find a better stationing selection. Figure~\ref{fig:all_days} shows that the majority of dates, ($> 60\%$), can find the minimal objective cost within the first 100 iterations.
\begin{figure}[t]  
    \centering
    \includegraphics[width=0.90\linewidth]{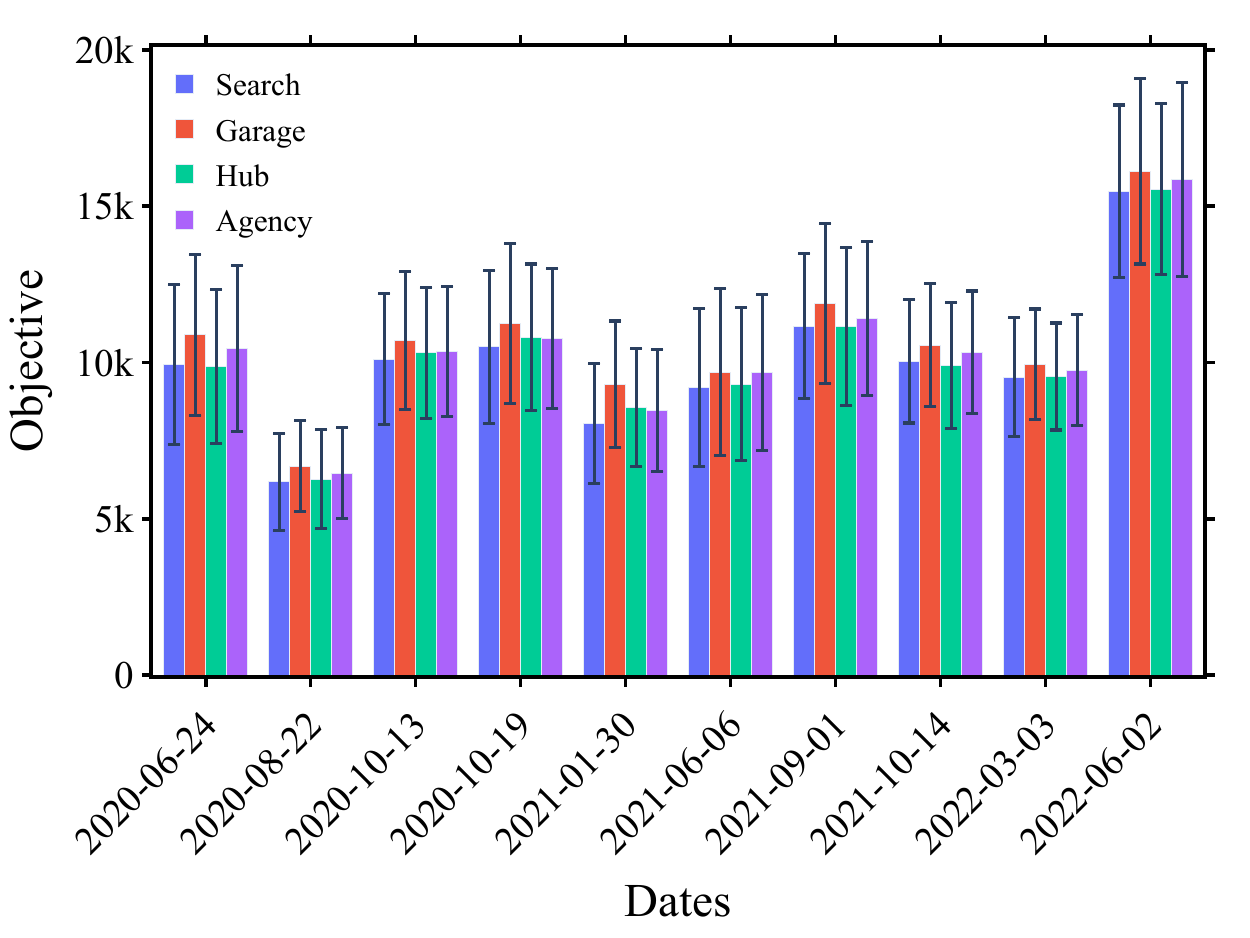}
    \caption{Comparison of different stationing plans across ten days. Lower is better.}
    \label{fig:search}
\end{figure}
Across multiple days, the stationing plans generated by the search algorithm result in much lower scores compared to the other stationing plans. Interestingly, having 5 buses stationed in the main hub can often outperform the searched stationing plan if there are no constraints on stationing. Finally, we compared whether optimizing once for multiple days would have better performance than running the optimizer every day. This is comparable to the agency running the algorithm a single time and then maintaining the same stationing across multiple days. Figure~\ref{fig:multi_day} shows that in almost all instances, optimizing for a single day will outperform a solution generated on multiple days.

\begin{figure}[t]  
    \centering
    \includegraphics[width=0.75\linewidth]{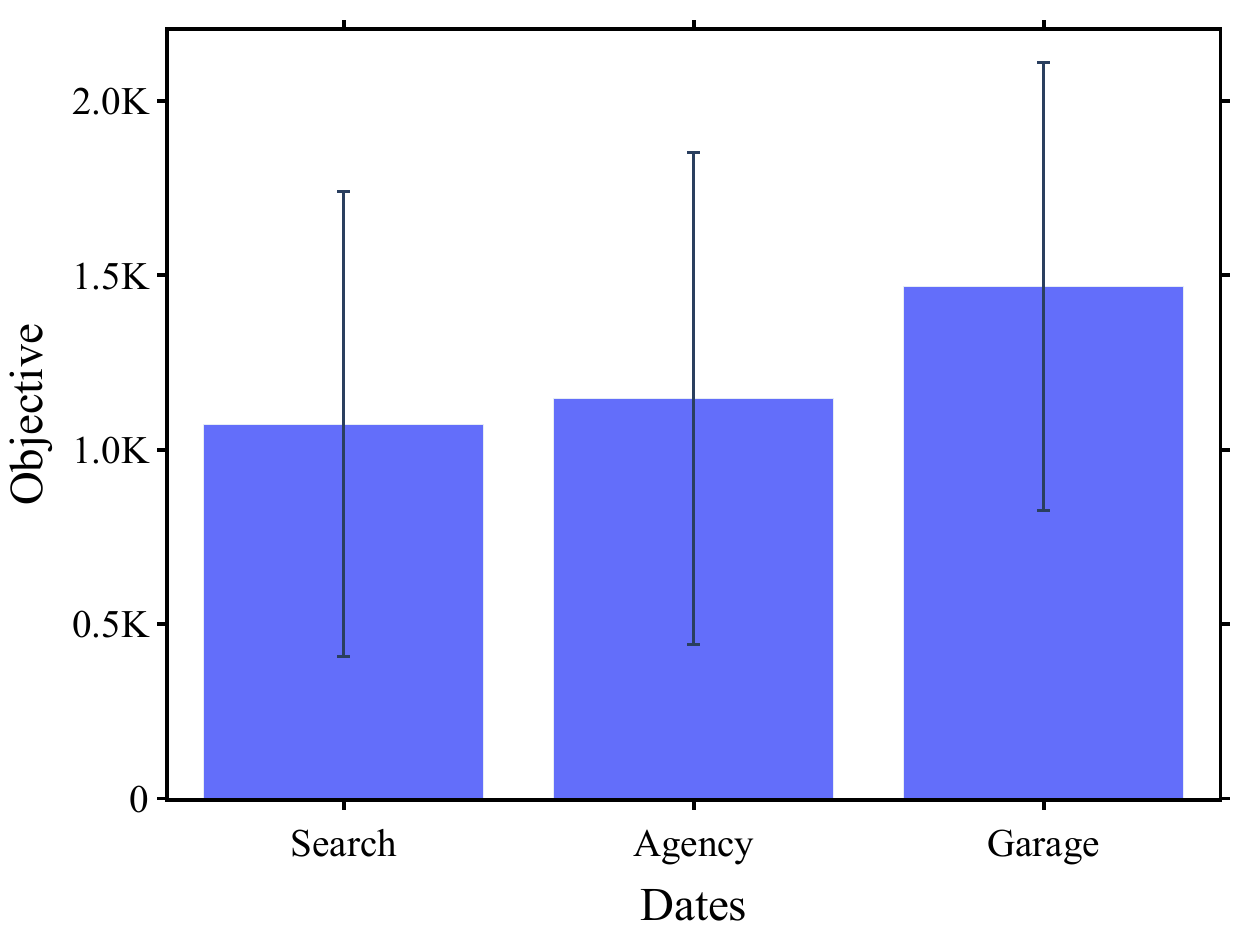}
    \caption{Using stationing plans from simulations using forecasted disruption and ridership on ground truth data. Lower is better.}
    \label{fig:realworld}
\end{figure}

\begin{table}[t]
\footnotesize
    \caption{Average cost of stationing plans across several days}
    \centering
    \begin{tabular}{|c||c|c|c|} \hline 
         \textbf{Stationing}& \textbf{Deadhead miles}&  \textbf{Deadhead minutes}&  \textbf{Passengers left}\\ \hline \hline
         Agency&  167&  175&  158\\ \hline 
         Hub&     \textbf{163}&  \textbf{170}&  158\\ \hline 
         Garage&  170&  179&  158\\ \hline 
         Search&  167&  175&  \textbf{132}\\ \hline 
    \end{tabular}
    \label{tab:stationing_results}
\end{table}

\paragraph{Stationing using forecasted disruptions} We use the stationing plans generated by the search algorithm on passenger forecast and disruption data and use it on the ground truth data. Disruptions are based on the actual events on which the forecasting model was trained. Figure~\ref{fig:realworld} shows that the current stationing plan used by the transit agency adapts well to day-to-day trips, including overages and possible disruptions, compared to simply stationing them at the garage or without any stationing plans. Meanwhile, our proposed stationing plan outperforms all other plans. This shows that the stationing plans based on forecast data are still able to perform better than baseline stationing plans. \cref{tab:stationing_results} shows that across several days, our approach can reduce the number of people left behind by 26 while matching both deadhead miles and minutes used by the current stationing plan used by the transit agency. This shows that by combining disruption forecasts and introducing an efficient heuristic algorithm, we can improve transit reliability with almost no cost on the part of the agency.

\section{Related Work}
\label{sec:related}

\iAron{decision-focused learning \url{https://guaguakai.github.io/IJCAI22-differentiable-optimization/static/images/IJCAI2022-decision-focused-learning-andrew-perrault.pdf}}

\noindent \textbf{Disruption forecasting} 
Public transit disruption forecasting has been a relatively under-researched area. 
Traditionally, the focus has been on predicting traffic accidents, computing delay times, or estimating passenger flow when disruptions occur, rather than forecasting the disruptions themselves. This has proven challenging due to the unpredictability of disruptions and the scarcity of data concerning them~\cite{vazirizade2021learning}. 
However, recent studies have shown that non-parametric machine learning models can be utilized to analyze extensive datasets comprising numerous independent variables to forecast traffic accidents and disruptions with high accuracy~\cite{yan2022traffic,Parsa2019TowardSH}. For instance, Yan et al.~\cite{yan2022traffic} employed the random forest model to forecast traffic accidents in the USA with high accuracy. 
Similarly, Parsa et al.~\cite{Parsa2019TowardSH} utilized XGBoost to predict traffic disruptions with high accuracy. 
The utilization of these non-parametric models provides a further advantage by elucidating the variables that play a crucial role in the prediction of disruptions ~\cite{chen2020modeling}. Another study by Liu et al.~\cite{liu2020impacts} examines the potential of using automated collection data to comprehensively analyze unplanned disruption impacts. They propose a systematic approach to evaluate disruption impacts on system performance and individual responses in urban railway systems using automated fare collection (AFC) data.

\noindent \textbf{Substitute Bus Services:} Most of the work done in public transit using buses centers around headway control or bunching prevention. Bunching can be attributed to several factors such as traffic congestion or clustered passenger arrivals. Traditional solutions to this are schedule-based approaches and adaptive headway approaches~\cite{DAGANZO2009913}. Adaptive headway control introduces bus holding times at certain stops on a route to reduce headway gaps. \citeauthor{PETIT201968} shows how dynamic bus substitution strategies can be used to contain deviations in the schedule~\cite{PETIT201968}. \citeauthor{zhang_metro_2018} studied the optimal time to dispatch a substitute bus. They found that buses should only be dispatched if an event takes longer than a period of~time~\cite{zhang_metro_2018}.

\noindent \textbf{Ridership Prediction:} When considering route planning for public transit, agencies need to be aware of the ridership. Thus, this has been well studied~\cite{ayman2022neural}. The current state of the art uses non-parametric approaches that build upon the non-linear relationships between the input and output variables without any prior knowledge. Neural network models, particularly LSTM~\cite{ridership_lstm} are used to take advantage of temporal characteristics of travel demand while incorporating spatial information such as neighborhood and census tracts~\cite{Dill2013PredictingTR}.

\section{Conclusion}
\label{sec:concl}

In this work, we presented a thorough strategy for solving the crucial problem of transit disruptions.  
We have created new ways for transportation agencies to proactively reduce delays and crowding by creating data-driven statistical models that precisely assess the possibility of disruptions. Our multifaceted methodology enabled the development of specific strategies that address both the macro and micro levels of transit service. 
Moreover, our approach minimized the number of passengers left behind by the strategic stationing of additional vehicles. In this way, the efficiency of transit operations can be increased, and the passenger experience can be greatly enhanced.
The implications of this approach go beyond simple operational improvements. This work contributed to a more robust, responsive, and customer-focused public transportation system by setting the standard for predictive modeling for transit disruptions. Future research may look towards improving forecast performance even more, modifying it for other transit systems, and integrating it with real-world decision support settings. This attempt is a significant step towards a more dependable and effective public transit system, where disruptions are managed not just reactively but also in advance, making proactive mitigation crucial for contemporary urban mobility.

\begin{acks}
This material is based upon work sponsored by the National Science Foundation under Grant CNS-1952011, CNS-2238815, by the Federal Transit Administration, and by the Department of Energy under Award DE-EE0009212. 
Any opinions, findings, and conclusions or recommendations expressed in this material are those of the authors and do not necessarily reflect the views of the National Science Foundation, the Federal Transit Administration, or the Department of~Energy. Results presented in this paper were obtained using the Chameleon Testbed supported by the National Science Foundation.
\end{acks}

\balance

\bibliographystyle{ACM-Reference-Format}
\bibliography{main}

\end{document}